\documentclass[10pt,twocolumn,letterpaper]{article}

\usepackage{wacv}              

\usepackage{graphicx}
\usepackage{amsmath}
\usepackage{amssymb}
\usepackage{booktabs}

\usepackage{times}
\usepackage{epsfig}

\usepackage{multirow}
\usepackage{cite}
\usepackage[table]{xcolor}
\usepackage{bbding}
\usepackage{enumerate}
\usepackage{pifont}
\usepackage{wasysym}
\graphicspath{{figures/}}

\usepackage{float}

\usepackage{tabu}
\usepackage{url}

\usepackage[pagebackref,breaklinks,colorlinks]{hyperref}

\usepackage[capitalize]{cleveref}
\crefname{section}{Sec.}{Secs.}
\Crefname{section}{Section}{Sections}
\Crefname{table}{Table}{Tables}
\crefname{table}{Tab.}{Tabs.}

\def\wacvPaperID{2640} 
\def\confName{WACV}
\def\confYear{2024}

\begin{document}

\title{Learning to Augment: Hallucinating Data for Domain Generalized Segmentation}


\author{Qiyu Sun$^{1}$, Pavlo Melnyk$^{2}$, Michael Felsberg$^{2}$, Yang Tang$^{1}$\\
$^{1}$ East China University of Science and Technology, $^{2}$ Linköping University \\
{\tt\small y20190063@mail.ecust.edu.cn, \{pavlo.melnyk,michael.felsberg\}@liu.se, yangtang@ecust.edu.cn}
}

\maketitle

\begin{abstract}
   Domain generalized semantic segmentation (DGSS) is an essential but highly challenging task, in which the model is trained only on source data and any target data is not available. 
   Existing DGSS methods primarily standardize the feature distribution or utilize extra domain data for augmentation.
    However, the former sacrifices valuable information and the latter introduces domain biases. 
    Therefore, generating diverse-style source data without auxiliary data emerges as an attractive strategy.
    In light of this, we propose GAN-based feature augmentation (GBFA) that hallucinates stylized feature maps while preserving their semantic contents with a feature generator.
    The impressive generative capability of GANs enables GBFA to perform inter-channel and trainable feature synthesis in an end-to-end framework.   
    To enable learning GBFA, we introduce random image color augmentation (RICA), which adds a diverse range of variations to source images during training. 
    These augmented images are then passed through a feature extractor to obtain features tailored for GBFA training.
    Both GBFA and RICA operate exclusively within the source domain, eliminating the need for auxiliary datasets.    
    We conduct extensive experiments, and the generalization results from the synthetic GTAV and SYNTHIA to the real Cityscapes, BDDS, and Mapillary datasets show that our method achieves state-of-the-art performance in DGSS.

   
\end{abstract}


\section{Introduction}
\label{sec:intro}
The success of current deep learning-based semantic segmentation methods~\cite{long2015fully,chen2017deeplab,badrinarayanan2017segnet} comes at the price of large quantities of densely annotated images, which are extremely time-demanding and expensive to acquire~\cite{cordts2016cityscapes}. 
Synthetic images with semantic labels generated by computer graphics can alleviate this problem by providing a relatively low-cost solution~\cite{richter2016playing, ros2016synthia}. 
However, the segmentation accuracy of models trained on synthetic datasets usually degrades dramatically  when tested on real scenes due to the domain shift~\cite{ganin2015unsupervised,ganin2016domain}---training and test splits coming from different data distributions. 
To tackle this, numerous techniques have been proposed, including domain adaptation (DA) methods~\cite{hoffman2018cycada,ma2021coarse} and domain generalization (DG) approaches~\cite{pan2018two,yue2019domain,choi2021robustnet}. 
DA methods require a particular target domain for joint training with the source domain, which may be unattainable in real-world applications. 
Contrarily, DG approaches only need the source domain for training, rendering them more practical and promising solutions for semantic segmentation~\cite{balaji2018metareg,li2018learning}.
This motivates us to focus our work on synthetic-to-real, domain generalized semantic segmentation (DGSS)~\cite{peng2022semantic}.

\begin{figure}
\centering
  \begin{subfigure}{0.98\linewidth}
  \includegraphics[width=0.98\linewidth]{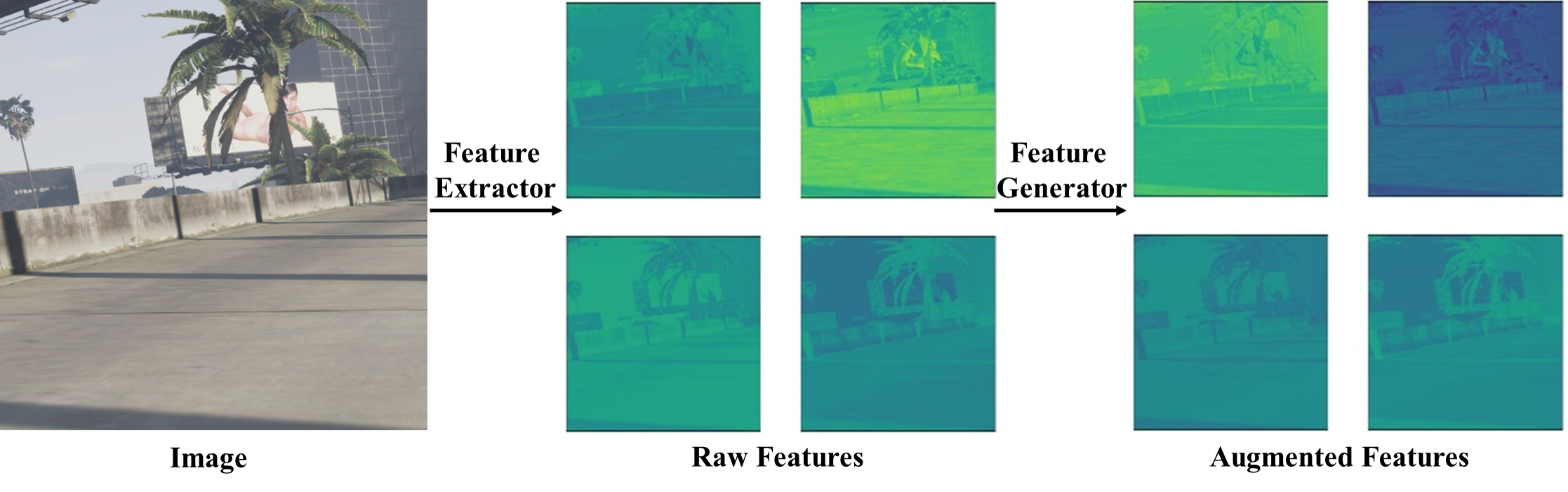}
    \caption{Illustration of learnable feature augmentation: Input image (left), raw features extracted from the image (middle), and augmented features generated by the feature generator from the raw features. The augmented features preserve the semantic context while introducing characteristic variations.}
    \label{intro_feat}
  \end{subfigure}

\begin{subfigure}{1\linewidth}
   \begin{minipage}{0.32\linewidth}
\centerline{\includegraphics[width=\textwidth]{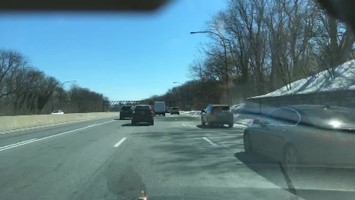}}
 \vspace{-6pt}
\centerline{\tiny{Target Images}}
\centerline{\includegraphics[width=\textwidth]{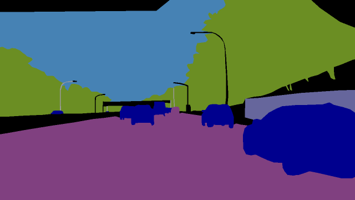}}
 \vspace{-6pt}
\centerline{\tiny{Ground Truth}}
\vspace{6pt}
\end{minipage}
 \begin{minipage}{0.32\linewidth}
 \centerline{\includegraphics[width=\textwidth]{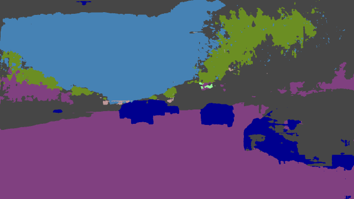}}
 \vspace{-6pt}
\centerline{\tiny{Our Baseline}}
\centerline{\includegraphics[width=\textwidth]{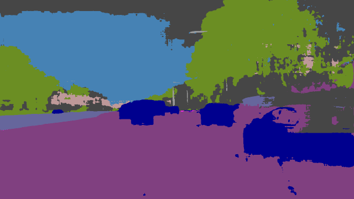}}
 \vspace{-6pt}
\centerline{\tiny{WildNet~\cite{lee2022wildnet}}}
\vspace{6pt}
\end{minipage}	
 \begin{minipage}{0.32\linewidth}
\centerline{\includegraphics[width=\textwidth]{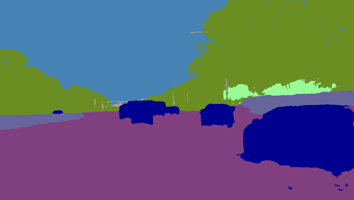}}
 \vspace{-6pt}
\centerline{\tiny{\textbf{Ours}}}
\centerline{\includegraphics[width=\textwidth]{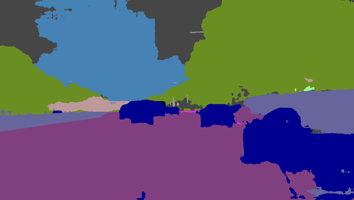}}
 \vspace{-6pt}
\centerline{\tiny{SHADE~\cite{zhao2022style}}}
\vspace{6pt}
\end{minipage}
\caption{Comparison of semantic segmentation results.}
\label{intro_result}
 \end{subfigure}

  \caption{The feature augmentation (a) and qualitative results (b). 
  Our method achieves superior generalization performance in the setting of DGSS but without additional data.}
\vspace{-10pt}
  \label{intro}
\end{figure}

Existing DGSS approaches either discard style information and standardize feature distribution to achieve domain-agnostic representations~\cite{pan2018two,pan2019switchable,choi2021robustnet,peng2022semantic}, or utilize auxiliary domain data to augment the source domain data to numerous styles~\cite{yue2019domain,huang2021fsdr,peng2021global,lee2022wildnet}. 
However, the former risks losing valuable information, while the latter introduces extra datasets during training, potentially leading to undesirable inductive biases~\cite{kim2022pin,peng2022semantic}. 
In light of this, generating stylized source data \textit{without additional training data}  is an appealing choice for DGSS. 
Previous work~\cite{zhao2022style} creates stylized data without extra data, which involves selecting basis styles and applying linear transformations through AdaIN~\cite{huang2017arbitrary} to adjust mean and standard deviation within individual feature channels.
However, due to its intra-channel, handcrafted feature augmentation, the diversity of produced styles remains limited. 
Recognizing the significance of diversifying data styles for DGSS, it becomes crucial to develop a method capable of generating a wider range of data styles. 

To address this, we introduce the GAN-based feature augmentation (GBFA) module, based on a network named FeatureGAN, to hallucinate stylized features from raw features while sustaining their semantic content, as illustrated in Figure~\ref{intro_feat}.
Replacing AdaIN with FeatureGAN enhances the feature augmentation capability as FeatureGAN
enables inter-channel and learnable feature augmentation in an end-to-end manner. 
This empowers the model to hallucinate a wider array of diverse data. 
To train the FeatureGAN, we introduce the random image color augmentation (RICA) module. 
Through RICA, we generate diverse-style images without requiring extra data. 
These augmented images, once processed by the feature extractor, give rise to features with varied distributions, which are then employed as the training data for the FeatureGAN.
RICA conducts image augmentation in the CIELAB color space~\cite{ibraheem2012understanding}, which better captures real dataset color variation compared to RGB.
The CIELAB space is designed to closely align with human perception, \eg, manipulating channel $A$ in CIELAB  controls color shifts like green to red, an option not available in RGB.
Similar RGB-to-CIELAB techniques are employed in DA methods~\cite{ma2021coarse, fantauzzo2022feddrive} to acquire source-content images with target style. Yet, these approaches demand specific target images and are confined to target domain style synthesis.
In contrast, RICA circumvents these restrictions, avoiding the necessity for explicit targets or auxiliary data, and facilitates the creation of a diverse array of styles for training GBFA or directly for DGSS. 




The proposed RICA and GBFA generate diverse images and features, providing complementary data augmentation without extra data, enabling segmentation networks to generalize to unseen target data and achieve state-of-the-art performance, as illustrated in Figure \ref{intro_result}.
The contributions of our work are summarized as follows:

\begin{itemize}
    \vspace{-4pt}
    \item [1)] We propose an effective RGB-to-CIELAB image color randomization mechanism named RICA for DGSS.  
    It eliminates the need for auxiliary datasets during training and relies only on the source data.

    \vspace{-4pt}
    \item [2)] We introduce the GBFA to perform inter-channel and learnable feature synthesis in an end-to-end manner, enabling the model to generate stylized features while maintaining their semantic contents.
    \vspace{-4pt}
    \item [3)] Extensive experiments conducted on benchmark datasets and multiple domain generalization tasks show that our proposed RICA and GBFA strategies operating in combination achieve state-of-the-art generalization results.
\end{itemize}

\section{Related Work}
In this section, we introduce related literature on synthetic-to-real semantic segmentation, including domain adaptation (DA) and domain generalization (DG).

\textbf{DA for semantic segmentation.}
 To bridge the gap between source and target domain, existing works prioritize DA techniques, which cater to a particular target domain by jointly learning from labeled source data and often-unlabeled target data.
Unsupervised DA (UDA) is the most attractive among different DA variants because the target domain is unlabeled in this setting, and hence, is less demanding. 
The mainstream UDA semantic segmentation methods can be divided into three categories: input-based (image translation)~\cite{hoffman2018cycada,ma2021coarse,kim2020learning}, feature-based (feature alignment)~\cite{hoffman2016fcns,luo2019significance,pan2020unsupervised}, and output-based (segmentation alignment)~\cite{tsai2018learning,vu2019advent,liu2021source}.
Even though DA increases the performance of the model on target domain to a large extent, it requires concurrent access to both source and target data, which is impractical and cannot be generalized to new target domains not seen during training.

\textbf{DG for semantic segmentation.} Recently, synthetic-to-real DGSS has begun to bring itself into purview due to its prospective real-world application prospect. 
In this line of work, the segmentation network is trained on synthetic domains and then tested on unseen real target domains.
Unlike DA approaches exploiting target domain data during training, DG methods utilize only the source domain and aim to generalize to multiple unseen target domains.
Existing DGSS methods can be divided into two groups~\cite{peng2022semantic}: 1) normalization and whitening and 2) domain randomization.
Additionally, some other works~\cite{kim2022pin, zhang2020generalizable} employ meta-learning-based methods for DGSS.

\begin{figure*}[t]
	\centering
	\includegraphics[width=1\textwidth]{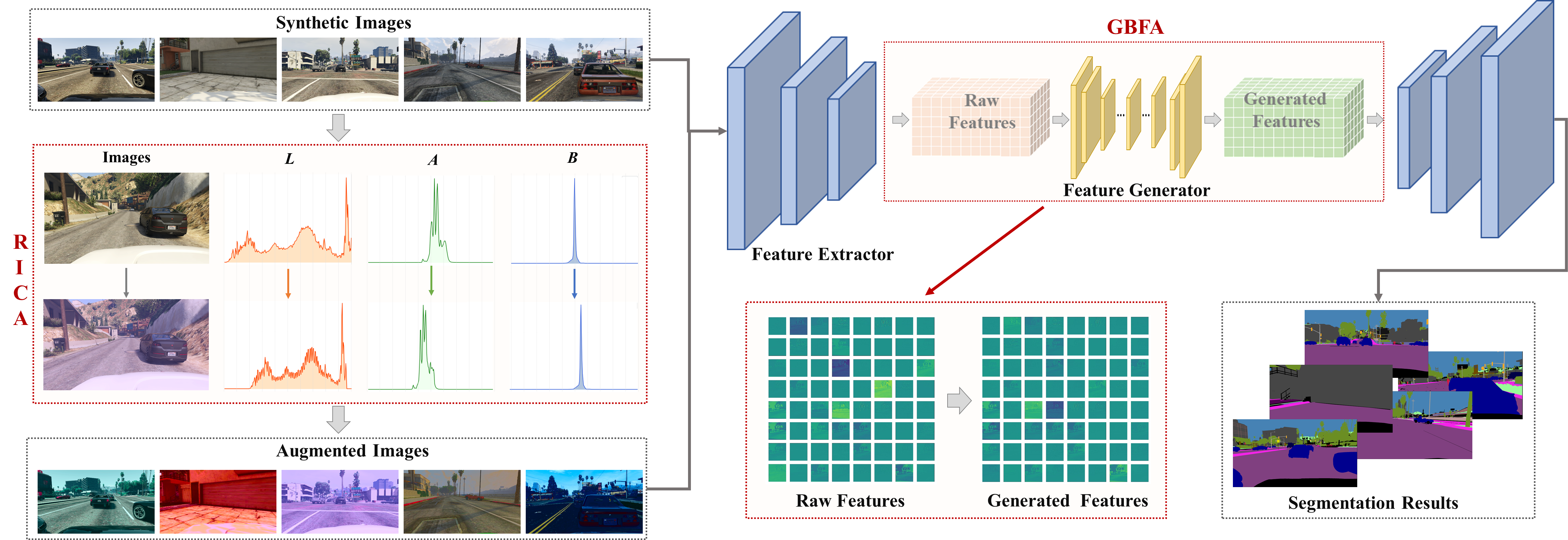}
	\caption{The training framework of our method: To generate new images with input-level data augmentation, we begin by inputting raw synthetic images into  RICA. Subsequently, both the original and augmented images are input into the segmentation model. The extracted features are further enhanced through our pre-trained Feature Generator (the generator of FeatureGAN) in the feature space.
 We utilize RICA exclusively during the training process, while we retain GBFA for inference purposes.
 }

	\label{main_framework}
 \vspace{-10pt}
\end{figure*}

Normalization and whitening methods utilize such techniques as instance normalization (IN)~\cite{ulyanov2017improved} or instance whitening (IW)~\cite{li2017universal} to standardize the feature distribution of different samples. 
Pan \textit{et al.}~\cite{pan2018two} propose a network architecture that is comprised of instance normalization (IN) and batch normalization (BN) to learn style-invariant features and preserve content information.
Choi \textit{et al.}~\cite{choi2021robustnet} suggest it is the style of images that causes domain shift, and thus disentangle features into style and content to selectively remove domain-specific style information.
Xu \textit{et al.}~\cite{xu2022dirl} employ feature sensitivity as the feature prior to enhancing the model's generalization capability.
On top of extracting domain-agnostic features, Peng \textit{et al.}~\cite{peng2022semantic} align category-level centers of different domains to obtain intra-category compactness.

Domain randomization methods heuristically explore the characteristics of potentially unpredictable target domains and try to generate multifarious styles of images for training from the source domain. 
Several works conduct data augmentation in image space~\cite{yue2019domain,huang2021fsdr,peng2021global} and some explore data manipulation in feature space~\cite{kim2021wedge,tjio2022adversarial,lee2022wildnet,zhao2022style}.
Some studies~\cite{yue2019domain,huang2021fsdr,lee2022wildnet} employ ImageNet~\cite{deng2009imagenet} as an auxiliary dataset for domain randomization. 
Other datasets are also utilized for domain randomization in some works~\cite{peng2021global,kim2021wedge,tjio2022adversarial}.
Moreover, Zhao \textit{et al.}~\cite{zhao2022style} diversify training samples by selecting basis styles from the source distribution and using AdaIN for intra-channel statistic adjustment, without relying on additional datasets as done by previous works.

Our method belongs to the domain randomization category and it stands out by not requiring additional datasets.
Unlike ~\cite{zhao2022style} which performs feature augmentation in separate channels by adjusting only the mean and standard deviation with AdaIN~\cite{huang2017arbitrary}, our method leverages the proposed FeatureGAN to hallucinate more diverse and informative features.

\section{Method}

In this section, we present two data-augmentation modules---RICA and GBFA for DGSS without using additional data. 
RICA aims to change the statistical distribution of the given source domain in image space, and GBFA improves the generalizability of the model by stylizing the features. 
Our main framework is shown in Figure \ref{main_framework}. 
First, the source images are fed into RICA for color augmentation. 
Then, the raw and augmented images are fed into our designed semantic segmentation model for training.
Unlike a traditional segmentation model pipeline, we integrate a feature generator into the model to perform inter-channel feature hallucination.

\subsection{Random Image Color Augmentation (RICA)}
\label{sec:RICA}

The vast majority of previous works~\cite{yue2019domain,huang2021fsdr,peng2021global,lee2022wildnet} rely on additional auxiliary domain data to transfer the source domain images to numerous styles.
On the contrary, our proposed RICA strategy aims to augment source-domain images using no auxiliary data.
Our main intuition is that RICA can deal with unseen target domains because it generates different kinds of statistical distributions, which prevent the segmentation model from over-fitting the characteristics of the given source data.
Consequently, the segmentation model can be more robust with respect to various data distributions across arbitrary unseen domains.

\begin{figure}[t]
	\centering
	\includegraphics[scale=0.37]{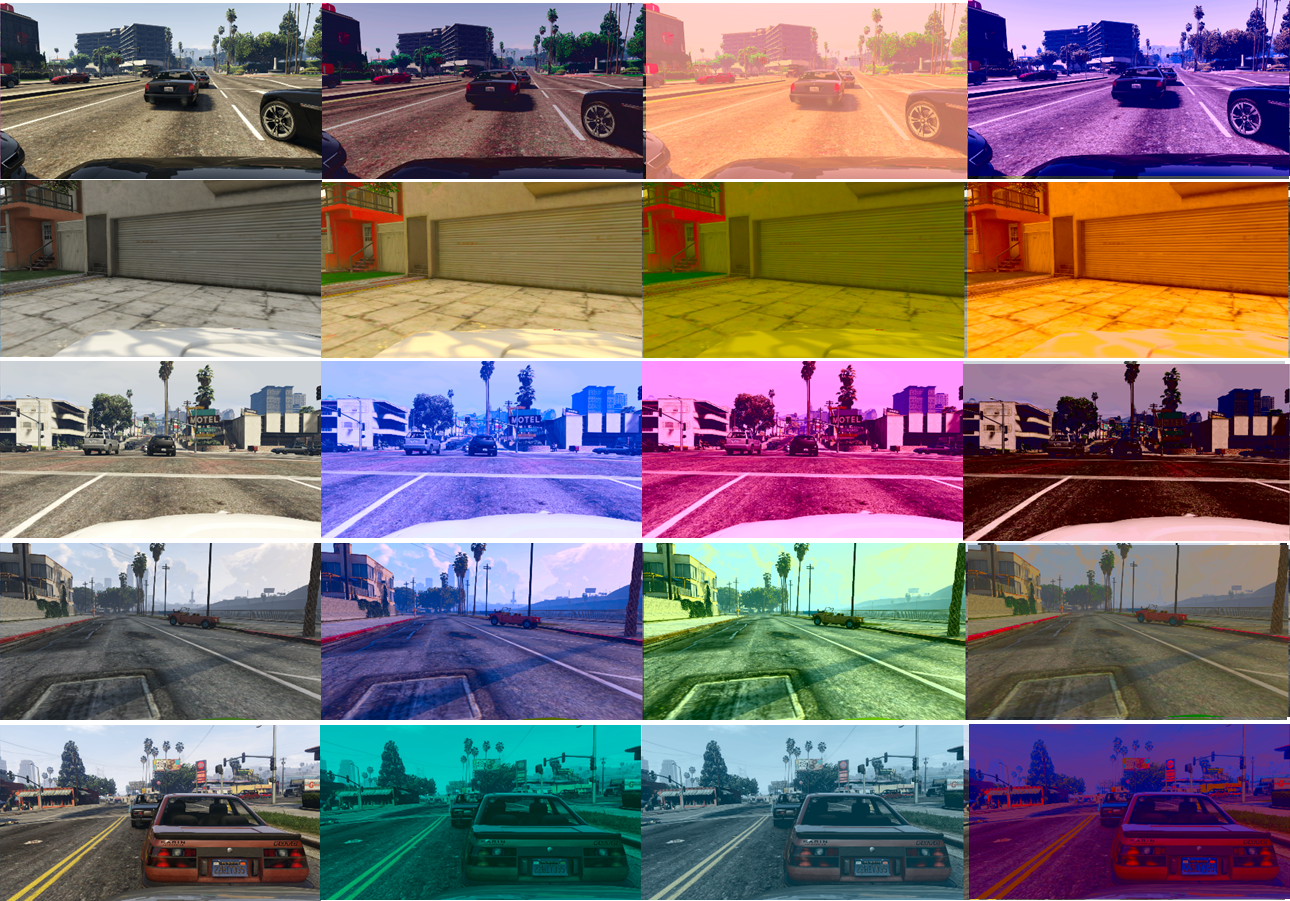}
	\caption{Color augmentation with RICA: Input images (left-most column) and RICA output (the remaining columns). 
 By adjusting the values in $LAB$ channels, RICA produces a range of images that retain their original semantic contents.}
	\label{lab_example}
 \vspace{-5pt}
\end{figure}

To conduct meaningful image augmentation, our idea is to use a color representation that is more intuitive for humans than the original RGB color model.
Therefore, we convert the images into the CIELAB model~\cite{ibraheem2012understanding} since it is designed to imitate human vision. 
The CIELAB color model contains three channels to express color: $L$ for perceptual lightness, $A$ and $B$ for four colors of human vision, depicting the shift from green to red and blue to yellow, respectively.
In other words, adjusting channel $L$ controls shifts from dark to light, channel $A$ controls shifts from green to red, and channel $B$ controls shifts from blue to yellow. However, attaining these effects is less straightforward in the RGB color model.

We use this human-perception-friendly color representation, because it allows us to augment the source data by randomly modifying the value of the three CIELAB channels, while preserving the semantic content of the original images.
We hypothesize that this augmentation can endow the training set with the different characteristics of various other datasets, improving the domain generalizability of the model.
To perform this augmentation, we first randomize the mean and standard deviation in the three channels, separately (RICA-Step1):
\begin{equation}
\label{adain}
\begin{aligned}
M^{c} = \sigma(M^{c}) \cdot \Big(\frac{R^{c}-\mu(R^{c})}{\sigma(R^{c})}\Big) + \mu(M^{c})\;,
\end{aligned}
\end{equation}
where $R^{c}$ is the raw color channels $c \in \{L, A, B\}$, $M^{c}$ is the corresponding color channels after the random augmentation of RICA-Step1.
$\mu(*)$ and $\sigma(*)$ represent the mean and standard deviation of each channel.
Then, we utilize the following linear mapping (RICA-Step2):
\begin{equation}
\label{rse1}
\begin{aligned}
N^{c} &= \frac{N^{c}_\mathrm{max}-N^{c}_\mathrm{min}}{M^{c}_\mathrm{max}-M^{c}_\mathrm{min}} \cdot (M^{c}-M^{c}_\mathrm{min}) + N^{c}_\mathrm{min}\\
&=\frac{M^{c}-M^{c}_\mathrm{min}}{M^{c}_\mathrm{max}-M^{c}_\mathrm{min}} \cdot (N^{c}_\mathrm{max}-N^{c}_\mathrm{min}) + N^{c}_\mathrm{min}\;,
\end{aligned}
\end{equation}
where $M^{c}_\mathrm{max}$ and $M^{c}_\mathrm{min}$ are the maximum and minimum values of the channel $M^{c}$,  $N^{c}_\mathrm{max}$ and $N^{c}_\mathrm{min}$ are the maximum and minimum values of the modified interval after RICA-Step2.

Let $S^{c} = N^{c}_\mathrm{max}-N^{c}_\mathrm{min}$ be the range of the modified interval and $T^{c} = N^{c}_\mathrm{min}$ be the starting point of the interval, then we obtain
\begin{equation}
\label{rse1}
\begin{aligned}
N^{c} = \frac{M^{c}-M^{c}_\mathrm{min}}{M^{c}_\mathrm{max}-M^{c}_\mathrm{min}} \cdot S^{c} + T^{c}\;.
\end{aligned}
\end{equation}
We randomly select the value of $\mu(M^{c})$, $\sigma(M^{c})$,  $S^{c}$ and $T^{c}$ for image color expansion in each channel  $c \in \{L, A, B\}$. 
The values of $M^c$ are clipped to [0, 255] in RICA-Step1, and the two steps of RICA work jointly to create more styles.


To sum up, we conduct diverse randomization in CIELAB color representation and then convert the images back to the RGB model for further training. 
In  RICA, we enlarge the range of each of the $LAB$ channels, which creates new, richer data distributions to train the segmentation model. 
As shown in Figure~\ref{main_framework},  RICA changes the characteristics of each channel and thus generates new images with different color styles.
More examples generated by  RICA are shown in Figure~\ref{lab_example}: as we can see, the image semantics are preserved, while the color is expanded.

\subsection{GAN-based Feature Augmentation (GBFA)}
\label{sec:trainingsteps}
 
We explore whether we can hallucinate multifarious distributions of features using GANs to diversify the characteristics of the features extracted from the source domain and thereby adapt to various other domains.
Thus, we design GBFA with FeatureGAN being its main component.

While the architecture of the FeatureGAN draws inspiration from the CycleGAN~\cite{zhu2017unpaired}, its core concept diverges significantly. The input and output of the FeatureGAN exclusively involve features, avoiding any image-level processing or reconstruction.
The FeatureGAN consists of two feature generators and two feature discriminators, that are trained simultaneously in a competitive process. 
We remove the first and last CNN layers of the generator in the CycleGAN to generate feature maps instead of images and keep the original discriminators.
The generators try to produce new features that are indistinguishable from real features, while the discriminators aim to differentiate between real and generated features. 
This dynamic and adversarial training nature empowers the generator to hallucinate stylistically diverse yet content-consistent features, as shown in Figure~\ref{intro_feat}, thus augmenting the training data for the segmentation network.
We refer to the part of the segmentation model from the input through the chosen layer for feature generator integration as the \textit{real\footnote{\textit{Real} as opposed to \textit{fake} in terms of the GANs training.}} feature extractor $F$, and the training samples for the FeatureGAN are generated by $F$.

The loss function of the FeatureGAN differs from the CycleGAN loss as we use the Kullback–Leibler (KL) divergence instead of the L1 loss for cycle consistency:
\begin{equation}
\footnotesize
\label{eq:gan_loss}
\begin{aligned}
\mathcal{L}_\textup{cyc}(G_{AB}, G_{BA})&=\mathbb{E}_{f_{A} \sim p_{\textup data}(f_{A})}[\textup{KL}(G_{BA}(G_{AB}(f_{A})) \vert \vert f_{A})] \\
&+\mathbb{E}_{f_{B} \sim p_{\rm data}(f_{B})}[\textup{KL}(G_{AB}(G_{BA}(f_{B}))\vert \vert f_{B})]\;,
\end{aligned}
\end{equation}
where $G_{AB}$ represents the feature generator that transforms features from domain A to B, and $G_{BA}$ generates in the opposite direction. $f_{A}$ and $f_{B}$ denote the real features extracted by $F$ from the respective domains.
We opt for the KL divergence because the L1 loss is typically employed to measure image differences, whereas the FeatureGAN generates features, making the KL divergence more suitable~\cite{zhang2021prototypical}.
The specific training steps are introduced in Section \ref{sec:training}.



\subsection{Training Steps}
\label{sec:training}

\begin{figure}
  \centering
  \begin{subfigure}{0.99\linewidth}
  \includegraphics[width=0.99\linewidth]{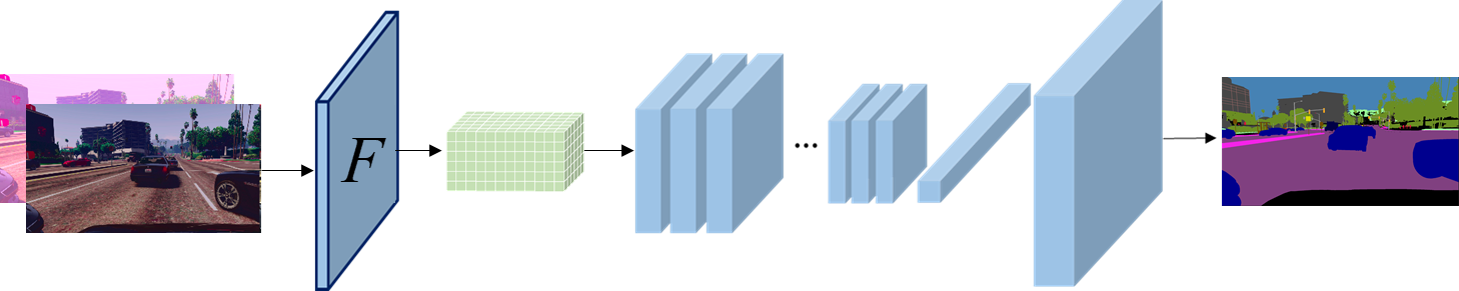}
    \caption{Step 1: The training of the real feature extractor $F$.}
    \label{step1}
  \end{subfigure}
  \hfill
   \begin{subfigure}{0.99\linewidth}
  \includegraphics[width=0.99\linewidth]{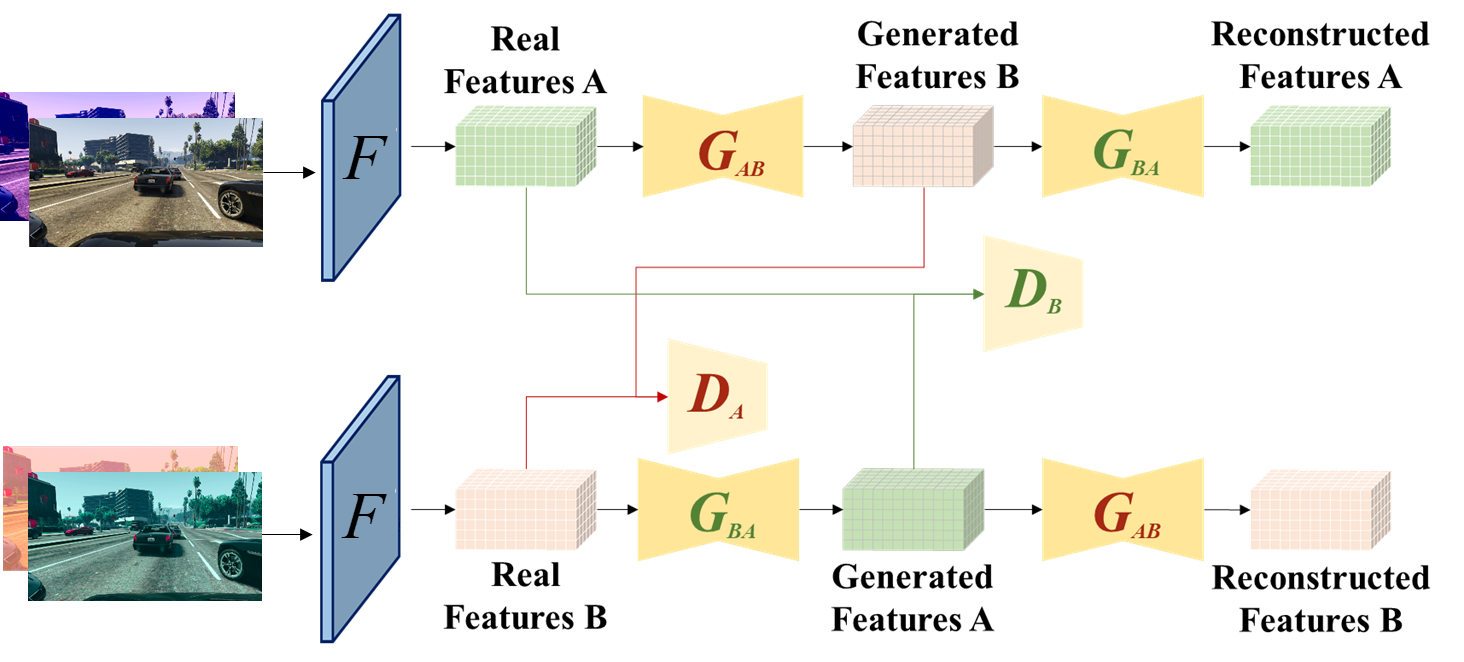}
    \caption{Step 2: The training of the FeatureGAN.}
    \label{step2}
  \end{subfigure}
  \hfill
    \begin{subfigure}{0.99\linewidth}
  \includegraphics[width=0.99\linewidth]{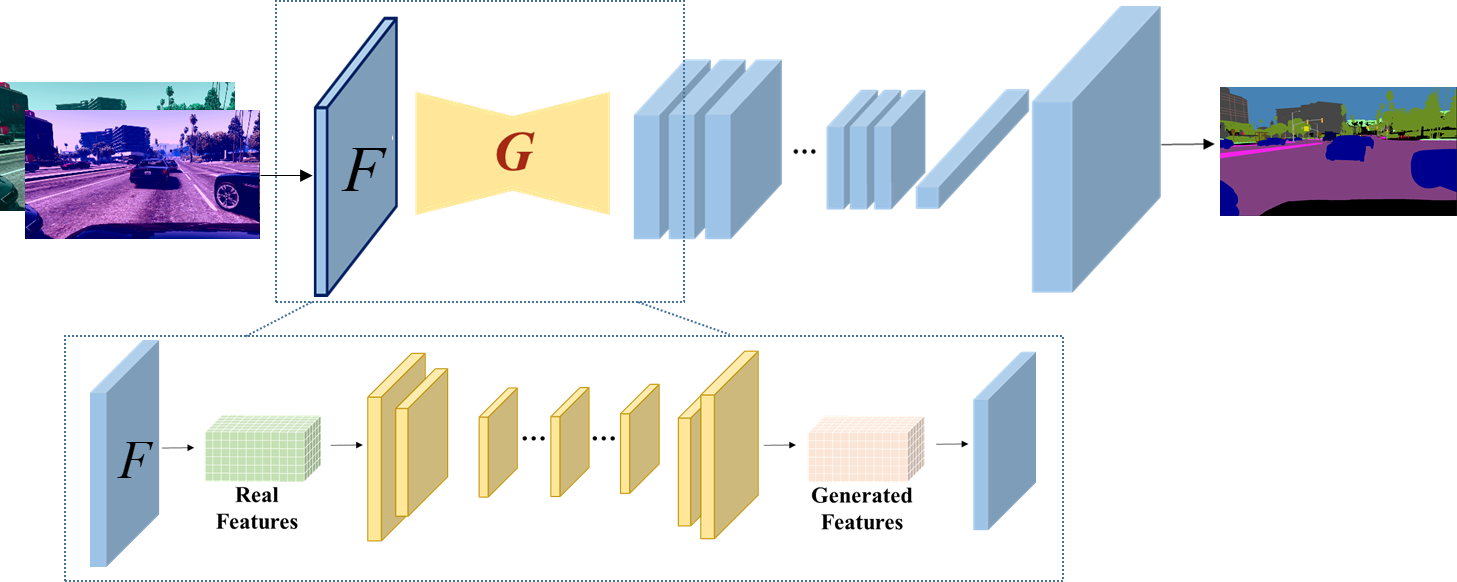}
    \caption{Step 3: The training of final segmentation model. We insert $G_{AB}$ into the final model as the feature generator $G$.}
    \label{step3}
    \vspace{-5pt}
  \end{subfigure}
  \caption{Training steps of our method, described in Section \ref{sec:training}.}
  \label{fig:RFDE}
 \vspace{-10pt} 
\end{figure}

The training of our method can be divided into three steps. 
First, we train the baseline segmentation model with images augmented by  RICA to obtain the real feature extractor $F$, as shown in Figure \ref{step1}. 
Second, we train the FeatureGAN using the $F$ obtained in Step 1 to generate features with distinct characteristics, illustrated in Figure \ref{step2}.
Finally, we plug the generator $G_{AB}$ of the FeatureGAN, further referred to simply as $G$, into the segmentation model in Step 1, as shown in Figure \ref{step3}, and perform the model training anew.

\textbf{Step 1: Train the real feature extractor $F$.}
We train the baseline semantic segmentation model with both raw and RICA-augmented images to obtain the real feature extractor $F$. 
Note that the generator $G$ can be inserted in the segmentation model after any chosen layer. 
If we insert $G$ after the first convolutional layer of the model, our $F$ is the first convolutional layer, as illustrated in Figure~\ref{fig:RFDE}.

\textbf{Step 2: Train the FeatureGAN.}
We fix the parameters of $F$ trained in Step 1 and utilize $F$ to generate real feature samples.
Since FeatureGAN aims to generate as many different styles of features as possible while maintaining their underlying semantic contents, the training samples (\ie, extracted features) should come from the same image but in different styles. 
Thus, we augment the source domain images using RICA, and then feed them into $F$ to obtain real samples to train the FeatureGAN. 
To be more specific, during the training process, we begin with a batch of raw images from a synthetic dataset. 
As illustrated in Figure~\ref{step2}, firstly, we augment these images twice with RICA, yielding two batches of images that share the same content but exhibit different styles. 
Next, each of these two batches is separately fed into the feature extractor $F$ to obtain real feature A and real feature B, which effectively generates features from two distinct domains. 
By repeating this process, we can generate a substantial amount of training data for FeatureGAN training.

\textbf{Step 3: Integrate  feature generator into the segmentation model.}
After FeatureGAN is trained, we fix the parameters of the generator $G_{AB}$ trained in Step 2 (simplified as $G$ in Step 3) and integrate it into the segmentation model, right after $F$, as shown in Figure \ref{step3}. 
Then, we train the new segmentation model with both raw images and the images augmented by  RICA. Just like in Step 1:
The output of $F$ is fed into $G$ which subsequently generates new, modified features, which are propagated into the following layers of the segmentation model for feature hallucination.


\section{Experiments}

\subsection{Experimental Setup}

\begin{table*}
\footnotesize

\centering

	\caption{Comparison of mIoU (\%) under DGSS setting using the backbone of ResNet-50 and ResNet-101. The source domain is GTAV (G), SYNTHIA (S), respectively, and the unseen target domains are  Cityscapes (C), BDDS (B), and Mapillary (M). The best and second best results are highlighted in \textbf{bold} and \underline{underline}. ``Auxiliary data" refers to any additional data used besides G and S. ``Baseline" represents the original basic model reported in the corresponding compared works.  ``/" means the result is not reported in the original paper.}
	\label{Quantitative_comparison}
 \begin{tabular}{ccccccccccc}
 
		\toprule[1pt]
		 \multirow{2}{*}{Backbone}&\multirow{2}{*}{Method}&\multirow{2}{*}{Auxiliary data} &\multicolumn{4}{c}{Train on GTAV (G)} &  \multicolumn{4}{c}{Train on SYNTHIA (S)}  \\
        \cmidrule(lr){4-7}\cmidrule(lr){8-11}
        
		{} &{}&{} & {$\rightarrow$ C} &{$\rightarrow$ B}  & {$\rightarrow$ M} & {Average}  &{$\rightarrow$ C }& {$\rightarrow$ B} &{$\rightarrow$ M} & {Average}   \\

		\toprule[1pt]



        
	& Baseline
        &\multirow{2}{*}{$\times$}
        &22.20  
        &/ 
        &/
        &/
        &/
        &/
        &/
        &/
        \\

        
	&IBN-Net~\cite{pan2018two} 
        &
        &29.60 
        &/
        &/ 
        &/ 
        &/ 
        &/
        &/
        &/ 
        \\

	\rowcolor{black!3}
        &Baseline 
        &
        &32.45  
        &26.73 
        &25.66
        &28.28 
        &28.36 
        &25.16
        &27.24
        &26.92 
        \\
		\rowcolor{black!3}
        &DRPC~\cite{yue2019domain}
        &\multirow{-2}{*}{$\checked$}
        &37.42  
        &32.14 
        &34.12
        &34.56
        &35.65
        &31.53
        &32.74
        &33.31
        \\
        &Baseline  
        &
        &28.95 
        &25.14
        &28.18 
        &27.42 
        &/ 
        &/
        &/
        &/ 
        \\
        &RobustNet~\cite{choi2021robustnet}
        &\multirow{-2}{*}{$\times$}
        &36.58
        &35.20
        &40.33
        &37.37
        &/
        &/
        &/
        &/
        \\
        \rowcolor{black!3}
        &Baseline          
        &
        &31.70 
        &/
        &/
        &/    
        &/
        &/
        &/
        &/
        \\
        \rowcolor{black!3}
        &GLTR~\cite{peng2021global}
        &\multirow{-2}{*}{$\checked$}
        &38.60 
        &/ 
        &/
        &/ 
       &/  
        &/
        &/
        &/ 
        \\
        &Baseline  
        &
        &29.32 
        &25.71
        &28.33 
        &27.79 
        &23.18 
        &24.50
        &21.79
        &23.16 
        \\
        &SAN-SAW~\cite{peng2022semantic}
        &\multirow{-2}{*}{$\times$}
        &39.75  
        &37.34
        &41.86
        &39.65
        &\underline{38.92}
        &\underline{35.24}
        &\underline{34.52}
        &\underline{36.23}
        \\

        \rowcolor{black!3}
        &Baseline 
        &
        &35.16 
        &29.71
        &31.29
        &32.05
        &/  
        &/
        &/
        &/ 
        \\
        
        \rowcolor{black!3}
        &WildNet~\cite{lee2022wildnet}
        &\multirow{-2}{*}{$\checked$}
        &{44.62}
        &{38.42}
        &\underline{46.09}
        &\underline{43.04}
        &/
        &/
        &/
        &/
        \\

        &Baseline 
        &
        &28.95 
        &25.14
        &28.18 
        &27.42 
        &/  
        &/
        &/
        &/ 
        \\
        
        &SHADE~\cite{zhao2022style}
        &\multirow{-2}{*}{$\times$}
        &\underline{44.65}
        &\underline{39.28}
        &{43.34}
        &{42.42}
        &/
        &/
        &/
        &/
        \\

        \rowcolor{black!3}
        &Baseline  
        &
        &33.29 
        &33.88
        &36.30 
        &34.49 
        &36.30 
        &26.14
        &30.47
        &30.97 
        \\
        
        \rowcolor{black!3}
        \multirow{-16}{*}{ResNet-50}
        &\textbf{Ours}
        &\multirow{-2}{*}{$\times$}
        &\textbf{47.82}  
        &\textbf{41.86}
        &\textbf{46.23}
        &\textbf{45.30}
        &\textbf{44.50}
        &\textbf{35.41}
        &\textbf{39.13}
        &\textbf{39.68} 
        \\
    
		\hline

	& Baseline 
        &
        &33.56
        &27.76  
        &28.33 
        &29.88
        &29.67 
        &25.64
        &28.73
        &28.01
        \\

		&DRPC~\cite{yue2019domain}        
        &\multirow{-2}{*}{$\checked$}
        &42.53 
        &38.72 
        &38.05
        &39.77
        &37.58
        &34.34
        &34.12
        &35.35
        \\

        \rowcolor{black!3}
        &Baseline  
        &\multirow{2}{*}{-}
        &33.40  
        &27.30
        &27.90 
        &29.53
        &/ 
        &/
        &/
        &/ 
        \\
        \rowcolor{black!3}
        &FSDR~\cite{huang2021fsdr}
        &\multirow{-2}{*}{$\checked$}
        &44.80
        &41.20
        &43.40
        &43.13
        &40.80
        &\textbf{37.40}
        &\underline{39.60}
        &\underline{39.27}
        \\

        &Baseline
        &
        &34.00
        &28.10
        &28.60 
        &30.23
        &30.20 
        &25.90
        &29.50
        &28.53 
        \\
        &GLTR~\cite{peng2021global}
        &\multirow{-2}{*}{$\checked$}
        &43.70
        &39.60
        &39.10
        &40.80
        &39.70
        &35.30
        &36.40
        &37.13
        \\

        \rowcolor{black!3}
        &Baseline  
        &\multirow{2}{*}{-}
        &30.64 
        &27.82
        &28.65 
        &29.04
        &23.85 
        &25.01
        &21.84
        &23.57
        \\
        
        \rowcolor{black!3}
        &SAN-SAW~\cite{peng2022semantic}
        &\multirow{-2}{*}{$\times$}
        &45.33
        &41.18
        &40.77
        &42.43
        &\underline{40.87}
        &35.98
        &{37.26}
        &38.04
        \\

        &Baseline  
        &
        &35.73 
        &34.06
        &33.42 
        &34.40
        &/
        &/
        &/
        &/ 
        \\
        
        &WildNet~\cite{lee2022wildnet}
        &\multirow{-2}{*}{$\checked$}
        &{45.79}
        &{41.73}
        &\textbf{47.08}
        &{44.87}
        &/
        &/
        &/
        &/
        \\

        \rowcolor{black!3}
        &Baseline 
        &
        &32.97 
        &30.77
        &30.68
        &31.47
        &/  
        &/
        &/
        &/ 
        \\
        
        \rowcolor{black!3}
        &SHADE~\cite{zhao2022style}
        &\multirow{-2}{*}{$\times$}
        &\underline{46.66}
        &\underline{43.66}
        &{45.50}
        &\underline{45.27}
        &/
        &/
        &/
        &/
        \\

        &Baseline  
        &
        &34.48 
        &35.78
        &38.26 
        &36.17
        &36.39
        &28.61
        &32.84
        &32.61
        \\
        \multirow{-14}{*}{ResNet-101}
        &\textbf{Ours}
        &\multirow{-2}{*}{$\times$}
        &\textbf{48.03}
        &\textbf{45.19}
        &\underline{46.26}
        &\textbf{46.49}
        &\textbf{44.99}
        &\underline{36.25}
        &\textbf{41.60}
        &\textbf{40.95}
        \\
		\hline

	\end{tabular}
  \vspace{-5pt}
\end{table*}

\subsubsection{Datasets}

In experiments, we utilize two synthetic datasets as source domains (GTAV~\cite{richter2016playing} and SYNTHIA~\cite{ros2016synthia}) and three  real datasets  as target domains (Cityscapes~\cite{cordts2016cityscapes}, Berkeley Deep Drive Segmentation (BDDS)~\cite{yu2020bdd100k}, and Mapillary~\cite{neuhold2017mapillary}). 
\textbf{Synthetic datasets:} 
GTAV is collected from the game Grand Theft Auto V  and contains 24,966 images. It is split into train, validation, and test sets, which consist of 12,403, 6,382, and 6,181 images, respectively. 
The subset of SYNTHIA, SYNTHIA-RAND-CITYSCAPES, contains 9,400 images and is split into 6,580 and 2,820 images for training and validation, following the convention in~\cite{choi2021robustnet, lee2022wildnet}. All of the synthetic datasets have pixel-wise semantic labels for training. 
\textbf{Real datasets:} 
In our experiments, we exclusively utilize the validation sets of Cityscapes, BDDS, and Mapillary for evaluation purposes, consisting of 500, 1,000, and 2,000 images, respectively.

\subsubsection{Implementation Details}

We use DeeplabV3+~\cite{chen2018encoder}\footnote{\url{https://github.com/VainF/DeepLabV3Plus-Pytorch}} as the semantic segmentation network in all the experiments. 
To conduct a fair comparison with existing methods~\cite{pan2018two,yue2019domain,huang2021fsdr,choi2021robustnet,peng2021global,peng2022semantic,lee2022wildnet,zhao2022style}, we adopt standard ResNet-50 and ResNet-101~\cite{he2016deep} as the feature extraction backbones.
The networks are trained with SGD as an optimizer~\cite{robbins1951stochastic} with
a weight decay of $10^{-4}$ and momentum of $0.9$.
The initial learning rates of the backbone and classifier are $2.5\cdot 10^{-3}$ and $2.5 \cdot 10^{-2}$, respectively, and they decrease with a power of 0.9 using the polynomial policy. 
The training batch size is set to 8 in the experiments.
During FeatureGAN training, two generators and two discriminators are trained simultaneously, with 11.364M and 2.828M parameters, respectively. Then, one generator is integrated into the final segmentation model, leading to an extra 11.364M parameters of the network.
Determined empirically, we randomly select $\mu(M^{c})$ from $0$ to $255$ for each channel, with $\sigma(M^{c})$ from $0$ to $100$,  $S^{c}$ from $30$ to $255$ for illumination channel $L$, and  $\sigma(M^{c})$ from $0$ to $15$,  $S^{c}$ from $30$ to $220$ for color channels $A$ and $B$ in  RICA. 


We choose the 19 categories that overlap with Cityscapes in both training and testing when utilizing GTAV as the source domain, while 16 categories are used when training on SYNTHIA only.
We evaluate the performance of our segmentation model using the mean intersection over union (mIoU) of the chosen categories.  
Unless otherwise stated, we insert the feature generator after the first convolutional layer to obtain the best results.
More implementation details are provided in the supplementary material.

\begin{figure*}[h]
 
 \centering
\begin{minipage}{0.032\linewidth}
 
 
\centerline{\includegraphics[width=\textwidth]{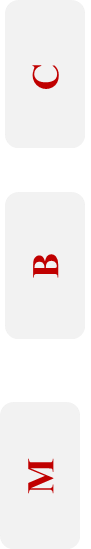}}
 
\centerline{ }
 
\end{minipage}
\begin{minipage}{0.13\linewidth}
 
 
\centerline{\includegraphics[width=\textwidth]{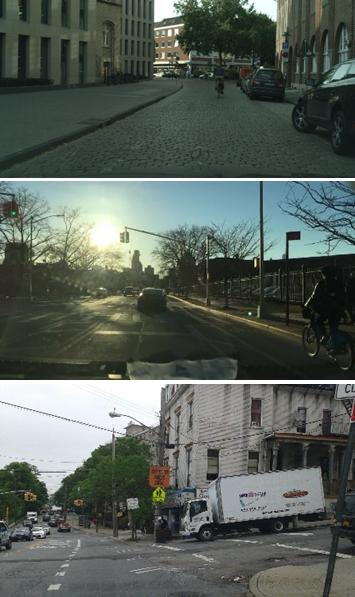}}
 \vspace{-3pt}
\centerline{\footnotesize{Target Images}}
  
\end{minipage}
\begin{minipage}{0.13\linewidth}
 
 
\centerline{\includegraphics[width=\textwidth]{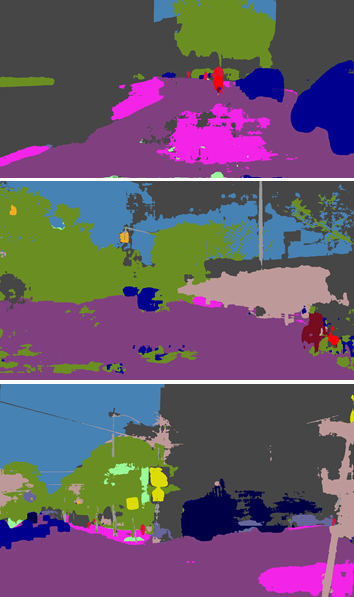}}
  \vspace{-3pt}
\centerline{\footnotesize{Our Baseline}}
 
\end{minipage}
\begin{minipage}{0.13\linewidth}
 
 
\centerline{\includegraphics[width=\textwidth]{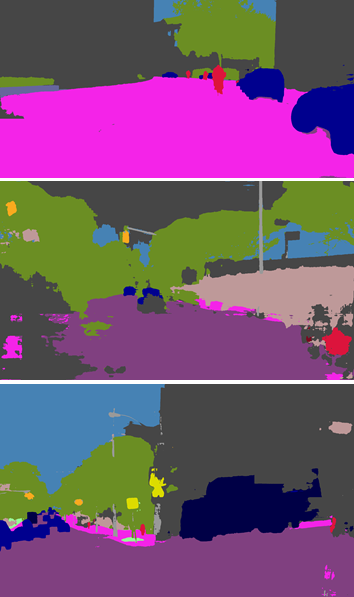}}
  \vspace{-3pt}
\centerline{\footnotesize{RobustNet~\cite{choi2021robustnet}}}
 
\end{minipage}
\begin{minipage}{0.13\linewidth}
 
 
\centerline{\includegraphics[width=\textwidth]{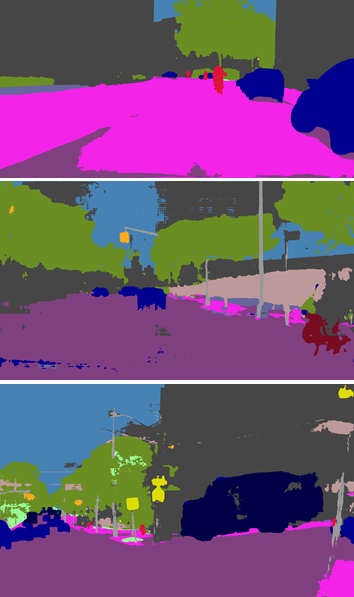}}
  \vspace{-3pt}
\centerline{\footnotesize{WildNet~\cite{lee2022wildnet}}}
 
\end{minipage}
\begin{minipage}{0.13\linewidth}
 
 
\centerline{\includegraphics[width=\textwidth]{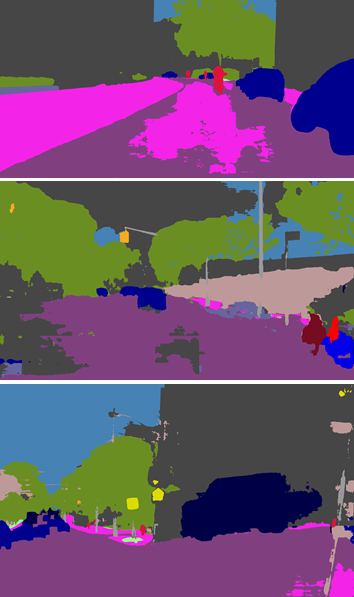}}
  \vspace{-3pt}
\centerline{\footnotesize{SHADE~\cite{zhao2022style}}}
 
\end{minipage}
\begin{minipage}{0.13\linewidth}
 
 
\centerline{\includegraphics[width=\textwidth]{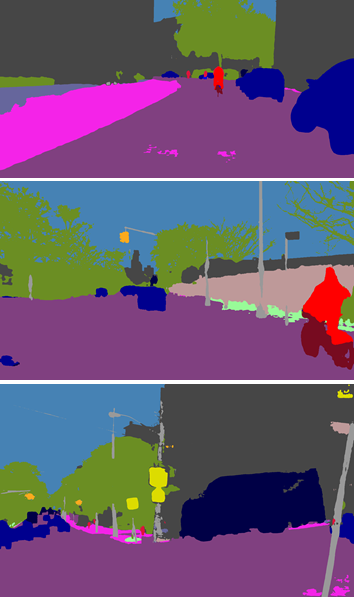}}
 \vspace{-3pt}
\centerline{\footnotesize{\textbf{Ours}}}
 
\end{minipage}
\begin{minipage}{0.13\linewidth}
 
 
\centerline{\includegraphics[width=\textwidth]{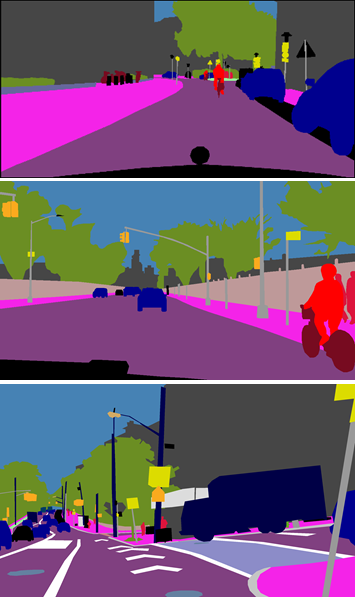}}
  \vspace{-3pt}
\centerline{\footnotesize{Ground Truth}}
 
\end{minipage}
%
\caption{Visual comparison of different models trained with ResNet-50 backbone (G$\rightarrow$ C, B, M). For a fair comparison, we utilize the models provided by the authors~\cite{choi2021robustnet,lee2022wildnet,zhao2022style} for visualization. 
Although three state-of-the-art methods (RobustNet~\cite{choi2021robustnet}, WildNet~\cite{lee2022wildnet}, and SHADE~\cite{zhao2022style}) have improved the average mIoU on three real datasets to 37.37$\%$, 43.04$\%$, and 42.42$\%$, their performance is still limited. Our method achieves superior generalization performance with an average mIoU of 45.30$\%$ by jointly synthesizing the color and feature maps, significantly outperforming the baseline.
More qualitative results on the three datasets are available in the supplementary material.
}
\label{visual_comparison}
 \vspace{-5pt}
\end{figure*}

\subsection{Results}
Quantitative comparisons are made between our method and the existing DGSS methods: IBN-Net~\cite{pan2018two}, DRPC~\cite{yue2019domain}, 
FSDR~\cite{huang2021fsdr}, RobustNet~\cite{choi2021robustnet},
GLTR~\cite{peng2021global}, SAN-SAW~\cite{peng2022semantic}, WildNet~\cite{lee2022wildnet}, and SHADE~\cite{zhao2022style}.  
We use GTAV and SYNTHIA separately as the source domain to train our model and then test it on the three unseen real-world datasets---Cityscapes, BDDS, and Mapillary. 
Table \ref{Quantitative_comparison} shows a comprehensive quantitative comparison of the semantic segmentation generalization performance of several models with ResNet-50 and ResNet-101 backbones. 
Our method achieves remarkably high generalization performance over multiple domains, while surpassing all the other methods using the ResNet-50 backbone, and being the best in all but two (G$\rightarrow$M and S$\rightarrow$B) comparisons for the ResNet-101 backbone.
To conceptualize the segmentation results, we present some qualitative results on each target domain in Figure~\ref{visual_comparison}. 
As we can see from the figure, our method can distinguish not only two easily-confused categories, \textit{sidewalk} and \textit{road} (row 1), but also various other objects, \eg, \textit{rider}, \textit{truck}, and \textit{traffic sign}. 
Furthermore, our method performs reliably in scenarios containing specular reflection (Figure~\ref{intro_result}) and low illumination (row 2).

We also conduct experiments of multi-domain generalization, which means training on both GTAV and SYNTHIA and then testing on unseen real domains.
Both the results of our method and the results available in related work are shown in Table \ref{multi_domain}.
Once again, our method demonstrates superior performance compared to the existing approaches.
Though the paper primarily aims to address the more practical synthetic-to-real generalization, we also present DGSS performance in real-to-synthetic and synthetic-to-synthetic scenarios in the supplementary material to further demonstrate the efficiency of the proposed method.

\begin{table}[t]
\centering
	\caption{Comparison of mIoU (\%) under the DGSS setting using the backbone of ResNet-50. The models are trained with multiple synthetic datasets (G + S $\rightarrow$ C, B, M).}
	\label{multi_domain}
 \resizebox{\linewidth}{!}{
	\begin{tabular}{ccccc}
		\toprule[1pt]
		Methods &  {G + S$\rightarrow$ C} &{G + S$\rightarrow$ B} &  {G + S$\rightarrow$ M} & Average \\
        \toprule[1pt]
         Baseline  & 35.46&25.09&31.94& 30.83\\ 
        RobustNet~\cite{choi2021robustnet} & 37.69&34.09&38.49&36.76\\ 			     
   
        \hline

        Baseline & 35.46&25.09&31.94& 30.83\\ 
        Kim \textit{et al.}~\cite{kim2022pin} & 44.51&38.07&42.70&41.76\\ 			     

        \hline

        Baseline & 35.46&25.09&31.94& 30.83\\ 
        SHADE~\cite{zhao2022style} & \textbf{47.43}&40.30&47.60&45.11\\ 	
        
        \hline

        Baseline  & 37.99&34.29&37.37&36.55\\ 
        Ours  & 47.36&\textbf{43.45}&\textbf{48.13}&\textbf{46.31}\\ 
   
        \toprule[1pt]
  
	\end{tabular}
 }
\end{table}

\subsection{Ablation Studies}
We conduct ablation experiments using the ResNet-50 backbone, generalizing from GTAV to Cityscapes, BDDS, and Mapillary.
First, we evaluate the  effectiveness of our proposed RICA and GBFA modules to see how they influence the model performance. 
The results are shown in Table \ref{ablation_study}, where Baseline represents the original DeeplabV3+ semantic segmentation model. 

The baseline model overfits the source domain and shows poor capability on real datasets.
On the other hand, when only RICA is applied, the model achieves a remarkably higher average mIoU of {43.55}$\%$ on three datasets with a {+9.06}$\%$ improvement. 
When only GBFA is utilized, the model achieves an average mIoU of {39.87}$\%$ with a {+5.38}$\%$ improvement. 
The combination of the two modules further boosts the model performance, attaining an average mIoU of {45.30}$\%$ with a {+10.81}$\%$ improvement.
In the following, we conduct more detailed ablation experiments for our RICA and GBFA modules.

\begin{table}[t]
\centering
	\caption{Ablation studies on RICA and GBFA (G$\rightarrow$ C, B, M). ``Baseline" denotes the original DeeplabV3+ using ResNet-50 backbone. The results are reported with mIoU (\%).}
	\label{ablation_study}
 \resizebox{\linewidth}{!}{
	\begin{tabular}{ccccccc}
		\toprule[1pt]
		{Methods}&RICA & GBFA&{G$\rightarrow$ C }  & {G$\rightarrow$ B} &  {G$\rightarrow$ M} & {Average} \\
       
         \toprule[1pt]
         Baseline & $\times$&  $\times$&  33.29& 33.88& 36.30&  34.49\\
         \hline
        Baseline + RICA & $\checked$ & $\times$ &{45.19}&{41.50}&{43.95}& 
 {43.55}\\ 
         \hline                 
        Baseline + GBFA & $\times$& $\checked$ & 36.88 &41.00 &41.72 & 39.87 \\	
         \hline                 
       All (Ours) & $\checked$& $\checked$  & \textbf{47.82}& \textbf{41.86}& \textbf{46.23}& \textbf{45.30} \\				     
   
        \toprule[1pt]
  
	\end{tabular}
 }
\end{table}

\textbf{RICA in different channels.} Table \ref{ablation_study_LAB} reports the influence of each $LAB$ channel on model generalization, along with the two randomization steps of RICA. 
Clearly, randomization in all channels leads to improvements, and both steps contribute to this improvement.
We observe that the augmentations in color channels $A$ and $B$ yield superior results compared to that in illumination channel $L$. This could be because the GTAV dataset already contains diverse illumination conditions, whereas the color tone of GTAV images is relatively consistent~\cite{richter2016playing}.
Moreover, additional experiments are provided in the supplementary material:
1) We replicate the identical randomization process of RICA in the RGB color model to prove the advantages of the RGB-to-CIELAB strategy;
2) We examine the influence of hyperparameters, including $\mu(M^{c})$, $\sigma(M^{c})$, and $S^{c}$, to show that RICA is robust to changes of hyperparameters;
3) We employ RICA in a Transformer-based model for DGSS to prove its efficiency further.

\begin{table}[t]
\centering
	\caption{Ablation studies on RICA in ResNet-50 backbone (G$\rightarrow$ C, B, M). $L$, $A$, and $B$ denote conducting data augmentation only in a single channel. ``RICA-Step1" and ``RICA-Step2" denote the two randomization steps introduced in Section \ref{sec:RICA}. The reported results are mIoU (\%).}
	\label{ablation_study_LAB}
 \resizebox{\linewidth}{!}{
	\begin{tabular}{ccccccccc}
		\toprule[1pt]
		\multirow{2}{*}{Method} &\multicolumn{3}{c}{Channel}&\multirow{2}{*}{G$\rightarrow$ C }  & \multirow{2}{*}{G$\rightarrow$ B} &  \multirow{2}{*}{G$\rightarrow$ M}  &  \multirow{2}{*}{Average}\\
        \cline{2-4}
         &$L$ & $A$ & $B$ & & & &          \\
         \toprule[1pt]
        Baseline &$\times$ &$\times$ &$\times$ &33.29& 33.88& 36.30&  34.49\\        
         \hline                 
       Baseline + $L$ &$\checked$& $\times$&$\times$  &34.10&37.08&38.94& 36.71\\				
         \hline
        Baseline + $A$ &$\times$ &$\checked$& $\times$  &44.74&39.95&\textbf{43.95}& 42.88\\       
        \hline
       Baseline + $B$ & $\times$ &$\times$ &$\checked$ &44.89&41.13&43.45& 43.15 \\  

       \hline
       Baseline + RICA-Step1& $\checked$&$\checked$ &$\checked$ &44.13&40.41&43.62& 42.72 \\  

       \hline
       Baseline + RICA-Step2& $\checked$&$\checked$ &$\checked$ & \textbf{45.40} & 38.29 & 42.42& 42.04 \\  
       
        \hline
     Baseline + RICA & $\checked$&$\checked$ &$\checked$ &{45.19}&\textbf{41.50}&\textbf{43.95}& 
 \textbf{43.55}\\ 
        \toprule[1pt]
  
	\end{tabular}
 }
\end{table}

\begin{table}[t]
\footnotesize
\centering
	\caption{Ablation studies on the position of  feature generator $G$ in ResNet-50 backbone (G$\rightarrow$ C, B, M).}
	\label{feature_GAN}
	\begin{tabular}{ccccc}
		\toprule[1pt]
		{Position of $G$}&{G$\rightarrow$ C }  & {G$\rightarrow$ B} &  {G$\rightarrow$ M} & {Average} \\
       
         \toprule[1pt]
         w/o $G$  & {45.19}&{41.50}&{43.95}& {43.55} \\
         \hline
        $G$ After Conv1 & \textbf{47.82}& \textbf{41.86}& \textbf{46.23}& \textbf{45.30}\\   
          \hline
       $G$ After Block Group1 &{46.34}&{41.67}&{43.60}&{43.87}\\   
         \hline                 
       $G$ After Block Group2 &46.47&39.60&44.40&43.49\\	
   
        \toprule[1pt]
  
	\end{tabular}
 \vspace{-10pt}
\end{table}

\begin{figure}[t]
	\centering
	\includegraphics[scale=0.25]{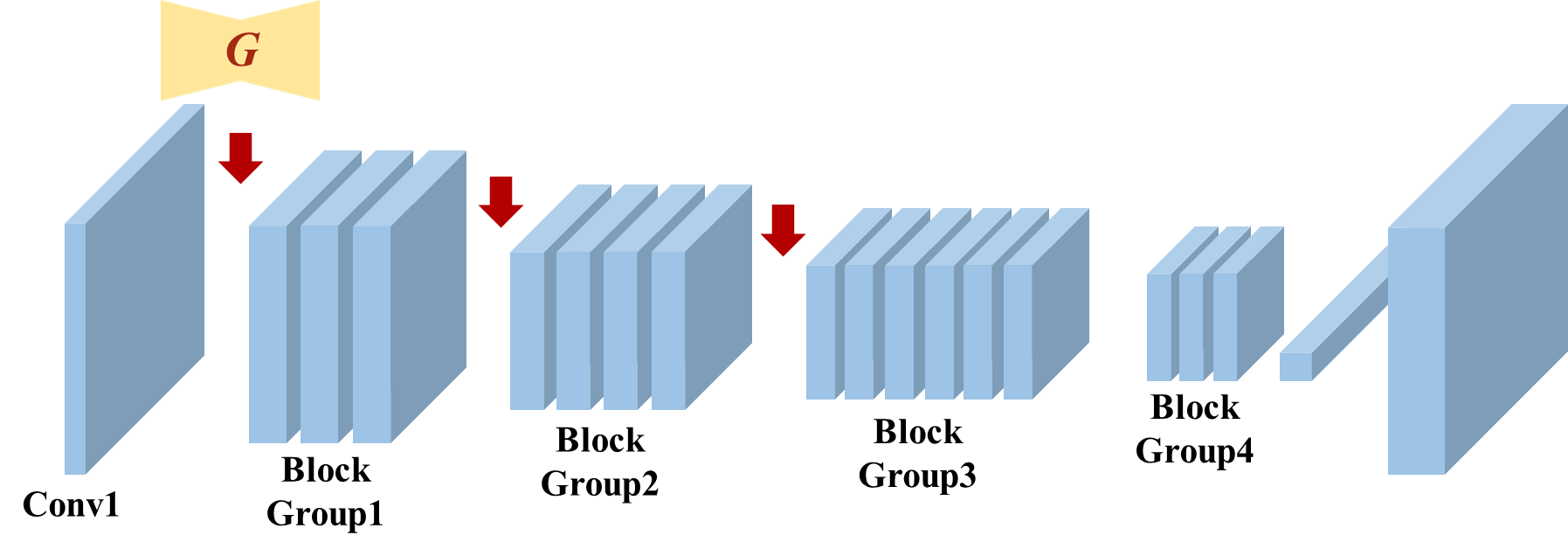}
	\caption{The illustration of the position of feature generator $G$.}
	\label{featgan_ablation}
\end{figure}

\textbf{GBFA in different layers.}
We also explore the impact of the position of the feature generator $G$ in the segmentation model, as illustrated in Figure \ref{featgan_ablation}. 
Namely, we use the features produced by the first convolutional layer, and then the first and second block groups for the respective FeatureGAN training. The results are shown in Table \ref{feature_GAN}.   
As we can see, placing GBFA deeper after Block Group 2 performs a little worse than the baseline.
It is explicable since deeper layers extract higher-level semantic information, and thus altering the high-level features can remove important details and edit semantic contents.
On the contrary, augmenting the first layer features preserves the valuable context information and vary style information, which  is beneficial for the segmentation task, as indicated by the higher performance of  RICA when placed after Conv1 (see Table~\ref{feature_GAN}).

\subsection{Further Analysis}

\begin{figure}
  \centering
  \begin{subfigure}{0.45\linewidth}
  \includegraphics[width=1\linewidth]{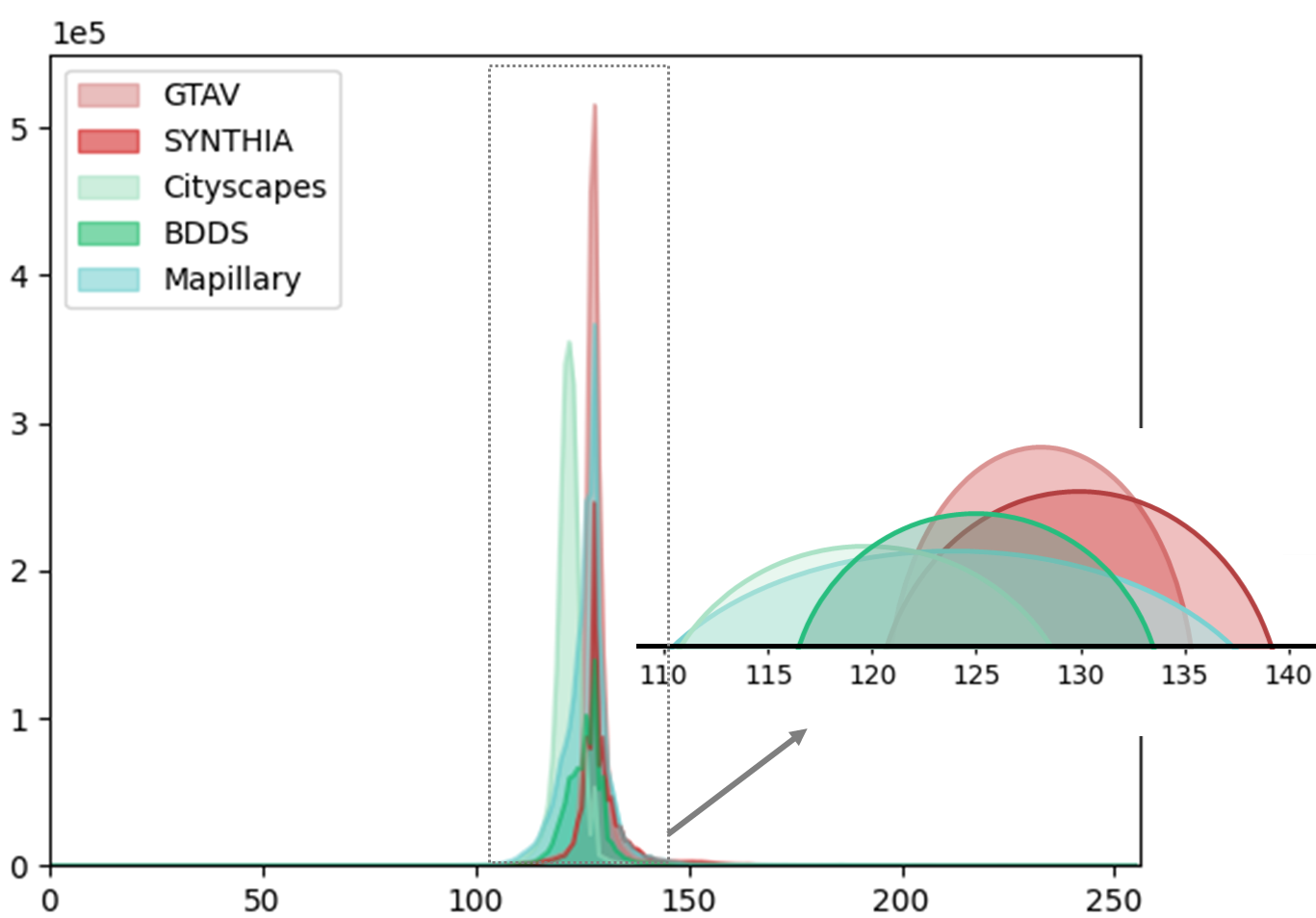}
    \caption{Before RICA}
    \label{before_lab}
  \end{subfigure}
  \hfill
   \begin{subfigure}{0.45\linewidth}
  \includegraphics[width=1\linewidth]{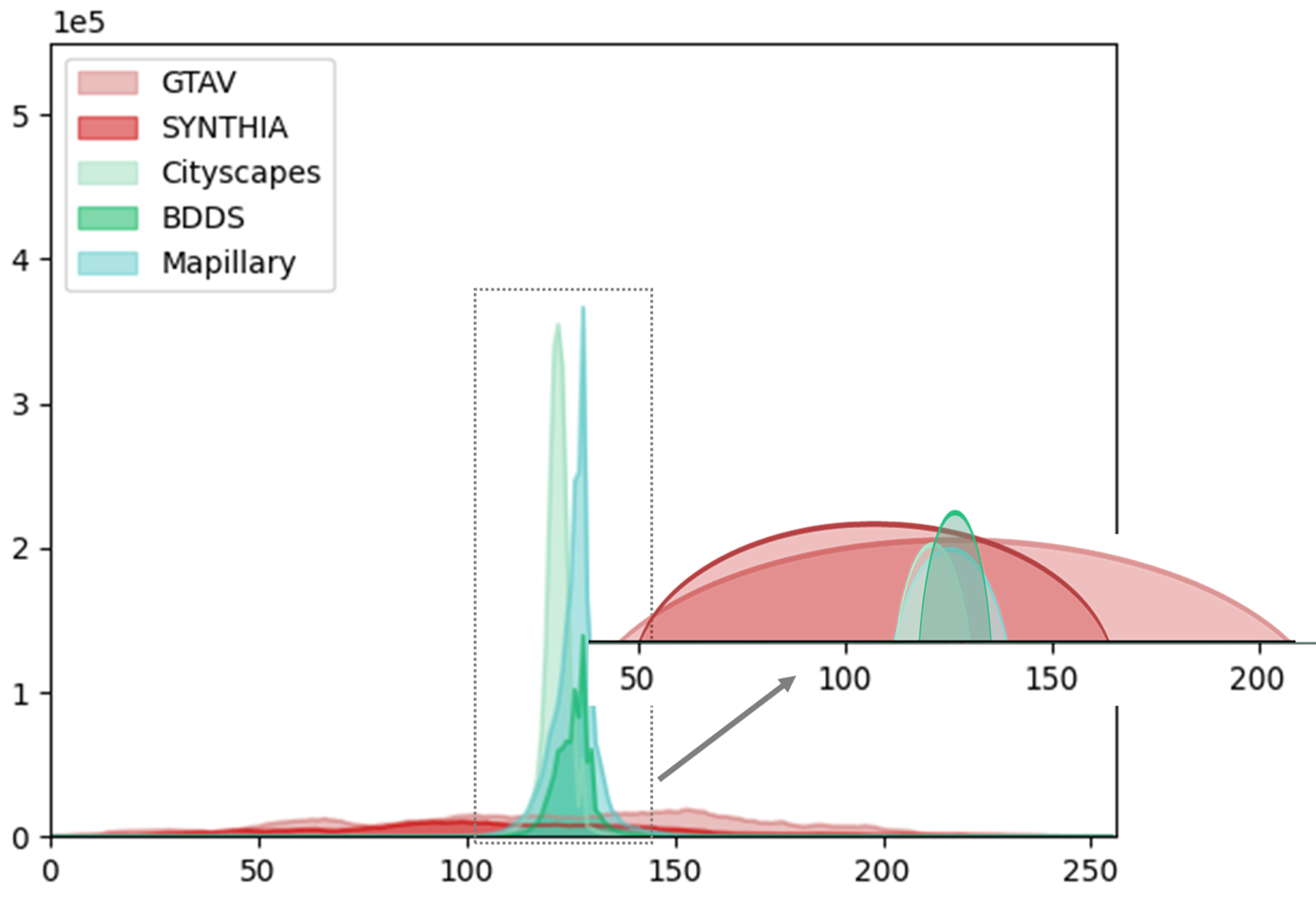}
    \caption{After RICA}
    \label{after_lab}
  \end{subfigure}
  \caption{Data distribution of different datasets in channel $A$. The zoomed-in regions are to display the overlap of the value ranges, and the relative height is arbitrary. After RICA, the value range of the synthetic data covers the whole distribution of the real data.}
  \label{different_dataset}
  \vspace{-10pt}
\end{figure}

\textbf{RICA.} To gain a better understanding of why RICA is so effective, we randomly select 100 images from each dataset and illustrate the distribution in channels $A$, as shown in Figure~\ref{different_dataset}. 
From Figure~\ref{before_lab}, the overlap of the value range between synthetic datasets (G and S in red) and real datasets (C, B, and M in green) is quite small.
We argue that this difference in data distributions, to some extent leads to the domain shift in semantic segmentation.
Since the value is concentrated in a small range, we locally zoom in on the range and visualize the distribution of the range in a more intuitive way.
As shown in Figure \ref{after_lab}, our method expands the channel value range of synthetic datasets to cover the whole distribution of real datasets.




\textbf{GBFA.} 
To prove our FeatureGAN indeed stylizes the features while preserving their semantic contents, we visualize four representative feature channels extracted by $F$ (real raw features) and the features generated by $G$ (generated augmented features) in Figure~\ref{intro_feat}. 
Here, $F$ corresponds to Conv1 as seen in Figure~\ref{featgan_ablation}.
We also provide a qualitative comparison of predictions with and without GBFA to show its effectiveness, which along with more visualizations of feature maps are shown in the supplementary materials.


\section{Conclusion}


In this paper, we generate diverse-style source data without auxiliary data for synthetic-to-real DGSS by the proposed RICA and GBFA. 
RICA introduces a diverse range of variations to source images and GBFA performs inter-channel and trainable feature hallucination.
Quantitative and qualitative results clearly indicate that our method exhibits notable generalization capabilities on different unseen target domains while using different backbones and source domains, achieving state-of-the-art results.
Several promising future directions to modify our method include exploring other generative models, \eg, the diffusion model, to simplify the training of GBFA, as well as altering the manual hyper-parameter selection in RICA with a learning-based module.


{\small
\bibliographystyle{ieee_fullname}
\bibliography{egbib}
}

\end{document}